\documentclass[conference]{IEEEtran}
\IEEEoverridecommandlockouts
\usepackage{graphicx}
\usepackage{arydshln}
\usepackage{amsmath}
\usepackage[dvipsnames]{xcolor}
\usepackage{multirow}
\usepackage{makecell}
\usepackage{adjustbox}
\usepackage{tabularx}
\usepackage{enumitem}
\usepackage{color}
\usepackage{siunitx}
\usepackage{amssymb}
\usepackage{tabulary,booktabs}
\usepackage{lipsum}
\usepackage{amssymb}
\usepackage{booktabs}
\usepackage{stfloats}
\usepackage{url}

\newcommand{\name}{{\texttt{FedCon}}}

\begin{document}

\title{FedCon: A Contrastive Framework for Federated Semi-Supervised Learning}

% \author{\IEEEauthorblockN{Under Review}}

\author{\IEEEauthorblockN{Zewei Long$^+$}\thanks{$+$ This work was done when the first author remotely worked at Penn State University.}
\IEEEauthorblockA{\textit{Department of CS} \\
\textit{UIUC}\\
Champaign, USA \\
zeweil2@illinois.edu}
\and
\IEEEauthorblockN{Jiaqi Wang}
\IEEEauthorblockA{\textit{College of IST} \\
\textit{Penn State University}\\
State College, USA \\
jqwangi@psu.edu}
\and
\IEEEauthorblockN{Yaqing Wang}
\IEEEauthorblockA{\textit{School of ECE} \\
\textit{Purdue University}\\
West Lafayette, USA \\
wang5075@purdue.edu}
\and
\IEEEauthorblockN{Houping Xiao}
\IEEEauthorblockA{\textit{Institute for Insight} \\
\textit{Georgia State University}\\
Atlanta, USA \\
hxiao@gsu.edu}
\and
\IEEEauthorblockN{Fenglong Ma}
\IEEEauthorblockA{\textit{College of IST} \\
\textit{Penn State University}\\
State College, USA \\
fenglong@psu.edu}
}

\maketitle

\begin{abstract}
Federated Semi-Supervised Learning (FedSSL) has gained rising attention from both academic and industrial researchers, due to its unique characteristics of co-training machine learning models with isolated yet unlabeled data. Most existing FedSSL methods focus on the classical scenario, i.e, the labeled and unlabeled data are stored at the client side. However, in real world applications, client users may not provide labels without any incentive. Thus, the scenario of labels at the server side is more practical. Since unlabeled data and labeled data are decoupled, most existing FedSSL approaches may fail to deal with such a scenario. To overcome this problem, in this paper, we propose {\name}, which introduces a new learning paradigm, i.e., contractive learning, to FedSSL. Experimental results on three datasets show that {\name} achieves the best performance with the contractive framework compared with state-of-the-art baselines under both IID and Non-IID settings. Besides, ablation studies demonstrate the characteristics of the proposed {\name} framework.
\end{abstract}


\section{Introduction}

% The next step for Artificial Intelligence (AI) should be the utilization of the fragmented, unlabeled data, which is common and waste in the real world. 

\begin{figure}[t]
\centering
\includegraphics[width=0.5\textwidth]{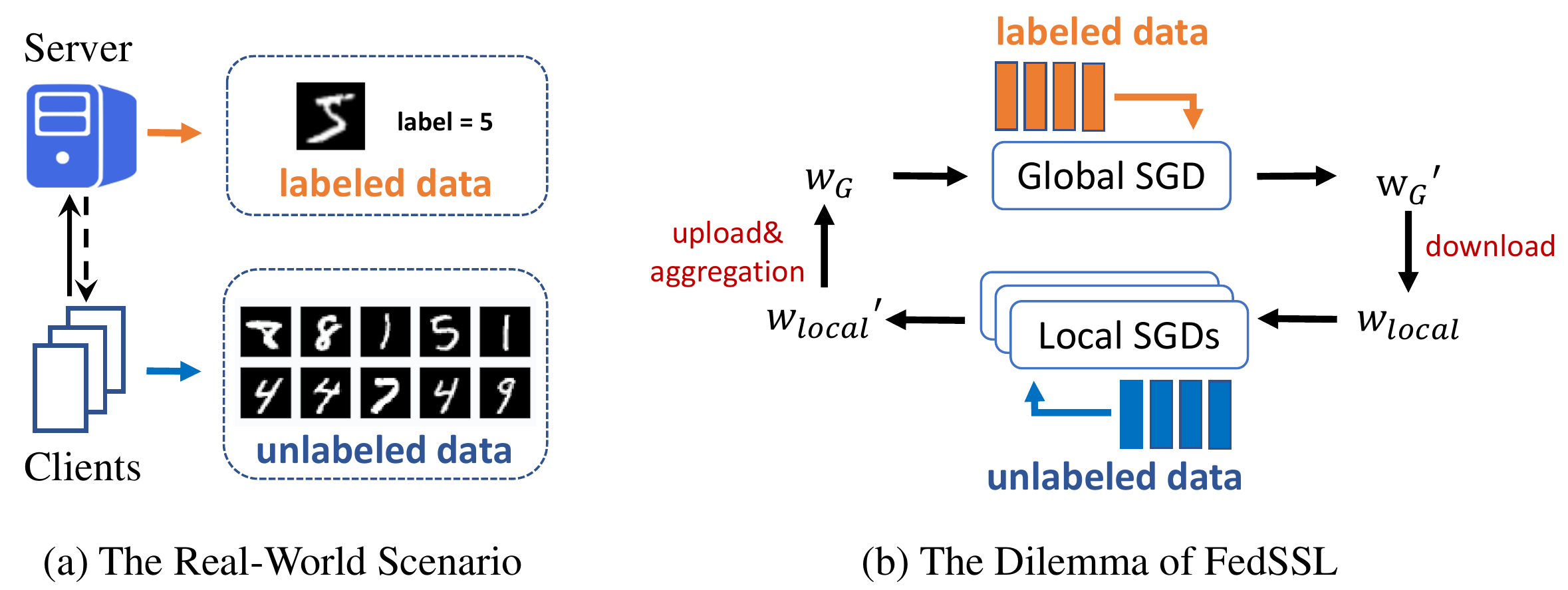} % Reduce the figure size so that it is slightly narrower than the column.
% \vspace{-0.2in}
\caption{Illustration of federated semi-supervised learning. (a) \emph{The real-world scenario}. The labeled data are only available at the server, while unlabeled data are available at the local clients. (b) \emph{The dilemma of FedSSL}.For each training round, the model cannot get access to unlabeled data at the clients and labeled data at the server simultaneously. In FedSSL, we firstly train the global model by global stochastic gradient descent (SGD) with labeled data at the server; then the local clients download the model and train with unlabeled data by local SGD; the local models are uploaded to the server and aggregated into a new global model for the next round. }

% For each training round, we firstly train the global model by global stochastic gradient descent (SGD) with labeled data at the server; then the local clients download the model and train with unlabeled data by local SGD; the local models are uploaded to the server and aggregated into the new global model for the next round. The model cannot get access to unlabeled and labeled data simultaneously during training.} 
\label{fig:two_settings}
\end{figure}

Federated Learning (FL), due to its unique characteristic of co-training machine learning models from fragmented data without leaking privacy~\cite{McMahan2017CommunicationEfficientLO,DBLP:journals/corr/abs-1902-04885,DBLP:journals/corr/abs-1912-04977}, has been widely applied in different applications~\cite{DBLP:journals/corr/abs-1811-03604,Yang2019FFDAF,Brisimi2018FederatedLO}. 
Most of existing FL studies~\cite{Sahu2018OnTC,Li2020OnTC,han2020robust} usually assume that the data stored in the local clients are fully annotated with ground-truth labels, but the server does not have any labeled data. 
However, this kind of assumption may be too strong for some real-world applications, since client users do not have enough incentives and efforts to label their generated data. Thus, a more practical and realistic scenario for federated learning is that the server holds all the labeled data, and clients have only unlabeled data as shown in Figure~\ref{fig:two_settings}. This scenario belongs to \emph{Federated Semi-Supervised Learning} (FedSSL).
Thus, how to utilize unlabeled data residing on local clients to learn the global model is a new challenge for FL. 

Recently, a few approaches~\cite{jin2020utilizing} are proposed to tackle this challenge by integrating classical semi-supervised learning techniques into the federated learning framework, such as FedSem~\cite{Albaseer2020ExploitingUD}, FedMatch~\cite{Jeong2020FederatedSL}, and SSFL~\cite{zhang2020benchmarking}. FedSem employs the pseudo-labeling to generate fake labels for unlabeled data based on the trained FedAvg~\cite{McMahan2017CommunicationEfficientLO} model with labeled data. FedMatch introduces a new inter-client consistency loss and decomposition of the parameters to learn the labeled and unlabeled data separately. SSFL discusses the gradient diversity in FL and utilizes group normalization (GN) and group averaging (GA) to improve the performance.
What these aforementioned methods have in common is that they focus on modifying semi-supervised methods to accommodate the new FedSSL setting. However, due to the decoupling of labeled data and unlabeled data, traditional semi-supervised methods will have a large performance loss~\cite{Tarvainen2017MeanTA}. Thus, it is urgent to design a new and specified semi-supervised learning framework for federated learning.

\begin{figure*}[t]
\centering
\includegraphics[width=0.8\textwidth]{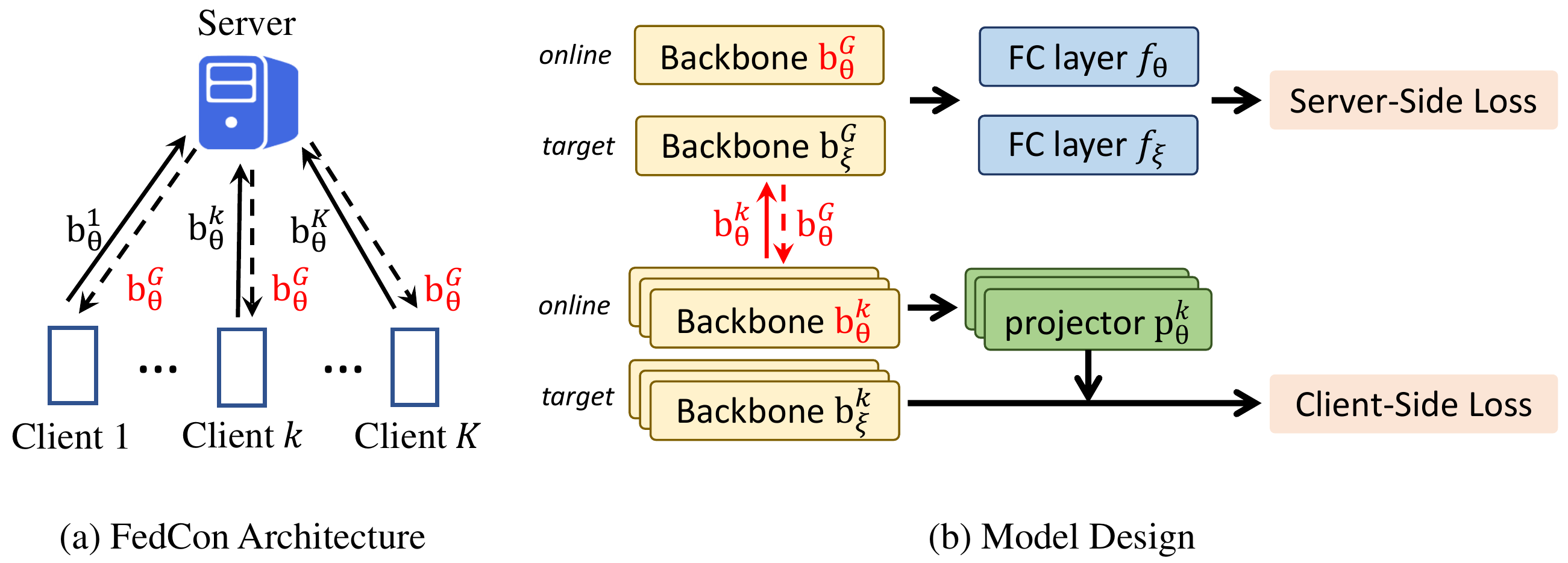} % Reduce the figure size so that it is slightly narrower than the column.
% \vspace{-0.1in}
\caption{Overview of the proposed {\name} Framework. (a) \textit{{\name} architecture}. Each client $k$ updates the contrastive network with its own data and uploads the model to the server. At the server side, the backbone of global model $b_\theta^G$ will be updated by averaging all the local parameters and further distributed to each client after the server update. (b) \textit{Model design}. The training of labeled data and unlabeled data shares most architectures except two output layers and loss functions.} 
\label{framework_overivew}
\end{figure*}

Towards this end, in this paper, we propose a novel and general federated semi-supervised learning framework, named {\name}, as shown in Figure~\ref{framework_overivew}. In particular, {\name} introduces a new \underline{\textbf{Con}}trastive network into \underline{\textbf{Fed}}erated learning to handle the challenge of the unlabeled data and further employs a unique two top-layer structure to solve the decoupling issue of labeled data and unlabeled data.

The contrastive network consists of two sub-networks: an online net and a target net. The online net keeps updating the parameters from the training data, and the target net updates slowly with the momentum mechanism and reserves the long-term information from the previous training. This division of labor enables the proposed {\name} framework to memorize the non-IID information and solve the data heterogeneity issue, which is one of key challenges in federated learning.

Except for utilizing a novel contrastive network, we also propose two different top-layer structures and loss functions for server-side and client-side updates. 
At the \textbf{server side}, we first train the model with labeled data. In particular, we use the designed contrastive network working with a backbone encoder $b$ to learn pairs of representations. Then a fully-connected (FC) layer is used to make predictions. We employ the cross-entropy classification loss and consistency loss at the server side to train the model. The learned parameters will distribute to each client.  
At the \textbf{client side}, we use the similar strategy to learn pairs of representations. However, since all the data are unlabeled, which limits us to use the previous two losses. Intuitively, if a pair of presentations is from the same input, they should be close to each other even in the projection space. This motives us to calculate the mean-squared-error loss for unlabeled data stored in each client. After training at the client side, the client parameters will be uploaded to the server.

In summary, the main contributions of this work are summarized as follows:
\begin{itemize}
    \item We recognize that the decoupling issue of labeled and unlabeled data may limit the power of existing federated semi-supervised learning approaches in real world scenarios.
    \item We propose a general, novel, and robust framework for federated semi-supervised learning, called {\name}\footnote{The source code of the proposed {\name} framework is publicly available at \url{https://anonymous.4open.science/r/fedcon-pytorch-E151} }, which handles the challenge of unlabeled data by introducing a contrastive network and two-output design. 
    \item  We show that {\name} outperforms state-of-the-art FedSSL baselines on three public datasets under both IID and non-IID scenarios. Moreover, ablation studies show that {\name} is more resilient to hyperparameter changes, which further demonstrates the great robustness of the proposed {\name}.

\end{itemize}

\section{Preliminaries}
In this section, we introduce some basic notations for FedSSL. FedSSL is a new collaborative learning paradigm, which aims to learn a global model from one server and several local clients using both labeled and unlabeled data. As mentioned in the Introduction section, in this paper, we focus on a more practical and realistic scenario -- the labeled data only store at the server side, and for all the clients, they don't have any labeled data.

\subsection{Input Data}
Let $\mathcal{D}_L = \{ (\mathbf{x}_1,y_1),\cdots, (\mathbf{x}_n,y_n) \}$ represent a set of labeled data stored in the server, where $\mathbf{x}_i$ ($i \in\{1, \cdots, n\}$) is a data instance, $y_i \in \{1, \cdots, C\}$ is the corresponding label, and $C$ is the number of label categories. For the $k$-th client, $\mathcal{D}^k_U = \{ (\mathbf{x}^k_1),\cdots, (\mathbf{x}^k_{n_k}) \}$ represents a set of unlabeled data, where $n_k$ is the number of unlabeled data. Note that the unlabeled data distributions at different clients/users may not follow the IID distribution, i.e., the Non-IID issue. 

% Unlike the supervised FL setting, the labeled data only exists in the central server. $\mathcal{D}_L = \{ (\mathbf{x}_1,y_1),\cdots, (\mathbf{x}_n,y_n) \}$ represents a set of labeled data, where $\mathbf{x}_i$ ($i \in\{1, \cdots, n\}$) is a data instance, $y_i \in \{1, \cdots, C\}$ is the corresponding label, and $C$ is the number of label categories. For the $k$-th client, $\mathcal{D}^k_U = \{ (\mathbf{x}^k_1),\cdots, (\mathbf{x}^k_n) \}$ represents a set of unlabeled data. Note that the unlabeled data distributions at different users might be non-iid. 

\subsection{Model Learning}
Similar to the common federated learning setting, the training process is done by multiple rounds of exchanging and updating model parameters between the server and clients. In each round of communication, we allow a small number of clients denoted as $B$ ($B \ll K$) to connect to the server and participate in the training process. Next, we demonstrate the FedSSL training pipeline, which can be slightly different from the standard FL setup. 

At the beginning of each round, the global model parameters $\boldsymbol{\theta}^G$ will be optimized by minimizing the loss function $\mathcal{L}_S(\mathcal{D}_L)$ with the labeled data $\mathcal{D}_L$ in the server. Then, each selected client such as $k$ will download the global model parameters $\boldsymbol{\theta}^G$ to train its local model parameters $\boldsymbol{\theta}^k$. In the $k$-th client, the local model parameters $\boldsymbol{\theta}^k$ will be updated by minimizing the loss function $\mathcal{L}_C(\mathcal{D}_U^k)$. After the local update, the local parameters $\{\boldsymbol{\theta}^1, \cdots, \boldsymbol{\theta}^B\}$ will be uploaded to the server for aggregating the new global model $\boldsymbol{\theta}^G$ in the next round. This procedure is repeated until $\boldsymbol{\theta}^G$ converges. 
%\jq{I am not sure it would better that we add a flow chart or something like that to illustrate?} \zw{fig 2(a) has illustrated the framework for {\name}, I think it's ok for us to demonstrate the general concept of fedssl by text?}
\section{{\name} Framework}

To alleviate the new dilemma caused by unlabeled data, we propose {\name}, an efficient FedSSL framework as shown in Figure~\ref{framework_overivew}. A contrastive network is employed as the general model to effectively handle the unlabeled data, which consists of two nets, i.e., an online net and a target net. We also design two different top layer structures and loss functions for server side and client side, respectively. Next, we will give the details of the proposed {\name} framework. 

\subsection{Contrastive network}

The goal of {\name} is to learn a model from both labeled and unlabeled data, which are distributed in the server and clients. Therefore, our model is built for (1) learning high-dimensional representations for classification tasks from unlabeled data and (2) learning classification model from labeled data based on the unlabeled-pretraining model. To achieve this goal, we employ the contrastive network to unite the global model and local models, which uses two subnetworks for model learning, i.e., the online and target nets. 
The online net keeps updating the parameters from the training data, and the target net updates slowly with the momentum mechanism and reserves the long-term information from the previous training.

The training process of the contrastive network is introduced as follows. For the online net $\theta$, we update its parameters by minimizing the loss with SGD.
Then we define the target net parameters $\xi_t$ at training step $t$ as the
exponential moving average (EMA) of successive $\theta_t$. Specifically, given a target decay rate $\alpha \in [0,1]$, after each training step we perform the following update, as:
\begin{equation}\label{ema}
\xi_t = \alpha \xi_{t-1}+ (1-\alpha) \theta_t.
\end{equation}

This unique design has two obvious benefits: (1) The existing of two networks enables the proposed {\name} framework to decrease the model diversity by adding the EMA. (2) The contrastive network promises the performance gain for training the unlabeled data at the clients and labeled data at the server. It unites the global model and local models with minimum cost.

\subsection{Server-side Design}

\begin{figure*}[!h]
\centering
\includegraphics[width=0.8\textwidth]{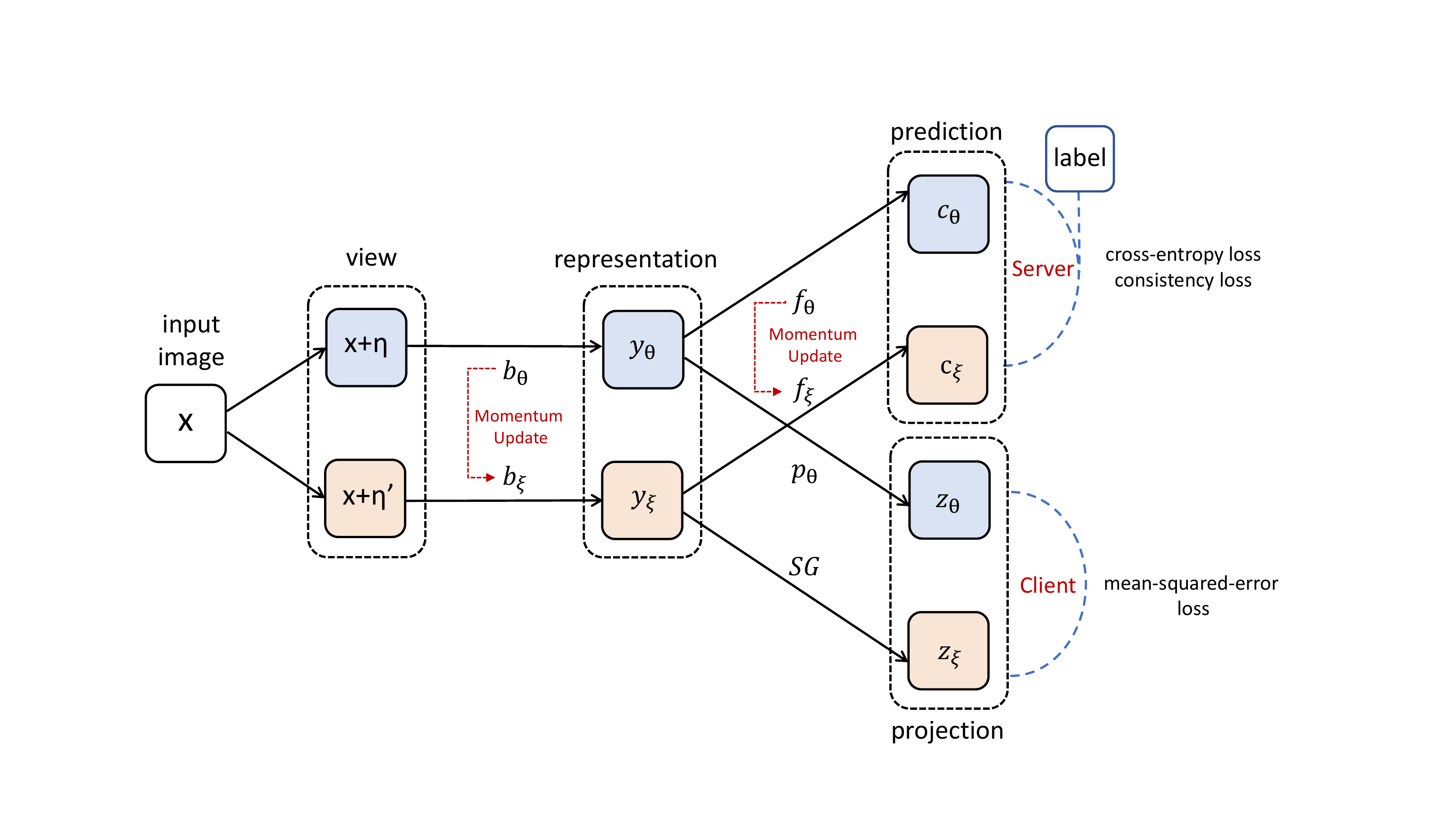} % Reduce the figure size so that it is slightly narrower than the column.
% \vspace{-0.1in}
\caption{ The model’s architecture. {\name} shares the same backbone $b_\theta$ \& $b_\xi$ at both server and clients. In the client side, {\name} minimizes a mean-squared-error loss between $z_\theta$ and $z_\xi$, where the online net $\theta$ are the trained net, the target net $\xi$ is updated by exponential moving average. In the server side, {\name} minimizes a cross-entropy loss and consistency loss between two model outputs and true label. ``SG'' is shorten for stop gradient.}
\label{siamese_network}
\end{figure*}

As mentioned in the previous subsection, we use the contrastive network to extract the information from the server data. Given an input image $x_i$, {\name} produces two new views $x_i+\eta$ and $x_i+\eta^{\prime}$ from $x_i$ by applying image augmentation techniques. From the first augmented view $x_i+\eta$, the online net outputs a representation $y_\theta \triangleq b_\theta(x_i+\eta)$ and a prediction $c_\theta \triangleq f_\theta(y_\theta)$, where $b_\theta$ is the backbone such as convolution layers, and $f_\theta$ is the FC layer. The target network outputs $y_\xi \triangleq b_\theta(x_i+\eta^\prime)$ and the target prediction $c_\xi \triangleq f_\xi(y_\theta^\prime)$ from the second augmented view $x+\eta^{\prime}$ as shown in Figure~\ref{siamese_network}.

We use the outputs of contrastive network, i.e., $c_\theta$ and $c_\xi$, and the ground-truth label $y_i$ to calculate the loss for the server data. We apply the widely used \textbf{cross-entropy loss} as the classification loss for the labeled data, i.e.,
\begin{equation}\label{classification_loss}
    \mathcal{L}_\theta \triangleq \frac{1}{n}\sum_{i=1}^{n} \sum_{c=1}^C p(y_i = c) \log f_\theta(b_\theta(x_i+\eta))
\end{equation}

Besides, we introduce the \textbf{consistency loss} into model training. The benefit of using this loss is to increase the model generalizability to deal with a more common scenario that the server has unlabeled data. In particular, given two perturbed inputs $\mathbf{x}_i + \eta$ and $\mathbf{x}_i + \eta^\prime$, the consistency loss disciplines the difference between the online net’s predicted probabilities $c_\theta \triangleq f_\theta(b_\theta(x_i+\eta))$ and the target net's predicted probabilities $c_\xi \triangleq f_\xi(b_\xi(x_i+\eta^\prime))$. The consistency loss is typically represented by the Mean Squared Error (MSE) as follows: 
\begin{equation}\label{mse_loss}
    \mathcal{J}_{\theta, \xi} \triangleq \frac{1}{n + m}\sum_{i = 1}^{n + m} \|c_\theta - c_\xi\|^2,
\end{equation}
where $n$ denotes the total number of labeled data, and $m$ denotes the total number of unlabeled data in the server.

We symmetrize the loss $L_\theta$ in Eq.~(\ref{classification_loss}) and $\mathcal{J}_{\theta, \xi}$ in Eq.~(\ref{mse_loss}) by separately feeding $x+\eta^\prime$ to the online net and $x+\eta$ to the target net to
compute $\mathcal{\tilde{L}}_{\xi}$ and $\mathcal{\tilde{J}}_{\theta, \xi}$.
%In each training step, we perform a stochastic optimization step to minimize $\mathcal{L}_S=(\mathcal{L}_\theta + \mathcal{L}_\xi )/2$ (or $\mathcal{L}_S=(\mathcal{L}_\theta + \mathcal{L}_\xi + \mathcal{J}_{\theta, \xi} + \mathcal{\tilde{J}}_{\theta, \xi})/2$ if there are unlabeled data in server) with respect to $\theta$ only, but not $\xi$. 
In each training step, we perform a stochastic optimization step to minimize $\mathcal{L}_S=(\mathcal{L}_\theta  + \mathcal{J}_{\theta, \xi} + \mathcal{\tilde{L}}_{\xi} + \mathcal{\tilde{J}}_{\theta, \xi})/2$ with respect to $\theta$ only.
The optimization of new $\xi$ is based on the moving average of $\theta$ and the current $\xi$ using Eq.~(\ref{ema}). The global model’s dynamics are summarized as follows:

\begin{equation}
\begin{array}{l}
\theta \leftarrow \text { optimizer }\left(\theta, \nabla_{\theta} \mathcal{L}_{S}, \eta\right), \\
\xi \leftarrow \tau \xi+(1-\tau) \theta,
\end{array}
\end{equation}
where $optimizer$ is an optimizer, and $\eta$ is the learning rate. %\zw{Eq 1 uses $\alpha$ as the EMA description for the general contrastive network, but server and clients use $\tau$ and $\mu$ to illustrate difference.}

After training with labeled data, the backbone $\boldsymbol{b}_\theta$ of the online net will be broadcast to selected clients while the fully-connected layer $\boldsymbol{f}_\theta$ is stored in the server. When the local training is finished, the backbones $\boldsymbol{b}_\theta$ will be uploaded to the server and aggregated into the global backbone, i.e. $\boldsymbol{b}_\theta^G \leftarrow \frac{1}{B} \sum_{k=1}^B \boldsymbol{b}_\theta^k$. The aggregated backbone and the fully-connected layer will be stitched together to form a new global online net $\theta$ for the next round learning.

\subsection{Client-side Design}

While the server-side model aims to learn the classification task, the goal of training client-side models is to learn the high-dimensional representations from the unlabeled data. Since the unlabeled data cannot provide the ground-truth labels as the labeled data do, there is no need to add an additional network architecture to learn the representations on the label space, i.e. $c_\theta$. Therefore, we only keep the backbone $\boldsymbol{b}_\theta$ of the global model and leave the fully-connected layer $\boldsymbol{f}_\theta$ in the server, which only focuses on label space representation learning. 

Similar to the server side, we output two representations $y_\theta \triangleq b_\theta(x_i+\eta)$ and $y_\xi \triangleq b_\theta(x_i+\eta^\prime)$ from two augmented views with the contrastive network. A naive solution is to calculate the consistency loss from these two representations. However, this loss can easily collapse into one constant solution without labeled data and cross-entropy loss. Therefore, as shown in Figure ~\ref{siamese_network}, a projector $\boldsymbol{p}_\theta$ is added to $\theta$, and the stop gradient (SG) is implemented in $\xi$ to encourage encoding more information within the online projection and avoid collapsed solutions as~\cite{grill2020bootstrap} does. To utilize the unlabeled data, we use the output of online projector $z_\theta \triangleq p_\theta(y_\theta)$ and the output of target backbone $z_\xi \triangleq y_\xi$ to calculate the loss, i.e., 

\begin{equation}\label{client_loss}
\mathcal{L}_{\theta, \xi} \triangleq\left\|z_\theta-z_{\xi}^{\prime}\right\|_{2}^{2}.
\end{equation}

We also symmetrize the loss $\mathcal{L}_{\theta, \xi}$ in Eq.~(\ref{client_loss}) by separately feeding $x+\eta^\prime$ to the online net and $x+\eta$ to the target net to
compute $\mathcal{\tilde{L}}_{\theta, \xi}$. At each training step, we perform a stochastic optimization step to minimize $\mathcal{L}_{C}=(\mathcal{L}_{\theta, \xi}+\mathcal{\tilde{L}}_{\theta, \xi})/2$ with respect to $\theta$ only, and then update $\xi$ based on $\theta$. The local model’s dynamics are summarized as

\begin{equation}
\begin{array}{l}
\theta \leftarrow \text { optimizer }\left(\theta, \nabla_{\theta} \mathcal{L}_{C}, \eta\right), \\
\xi \leftarrow \mu \xi+(1-\mu) \theta.
\end{array}
\end{equation}

After the training of local unlabeled data, the backbone $\boldsymbol{b}_\theta$ of the online work is uploaded to server while the projector $\boldsymbol{p}_\theta$ is stored in the client. When this client is selected for the next local training, the backbone $\boldsymbol{b}_\theta$ from the server and the projector $\boldsymbol{p}_\theta$ in the client will be combined to build the new local online net $\theta_k$.

\section{Experiments}

In this section, we evaluate the performance of the proposed {\name} framework, including both the IID and non-IID settings. We observe that the proposed {\name} achieves the best performance compared with all existing baselines. Furthermore, we conduct ablation experiments to analyze the efficiency of our model. Note that the main experiments focus on the scenario that there is no unlabeled data on the server. However, we also conduct a preliminary experimental analysis to validate the performance changes of the proposed {\name} in a more general scenario, that is, incorporating unlabeled data on the sever. The results are shown in Section~\ref{consostency_loss}.

\subsection{Experimental Setup}
%\jq{"IID" and "non-IID" writing should be consistent}
\noindent\textbf{Datasets}. 
In our experiments, we use three public datasets, including MNIST\footnote{\url{http://yann.lecun.com/exdb/mnist/}}, CIFAR-10\footnote{\url{https://www.cs.toronto.edu/~kriz/cifar.html}}, and SVHN\footnote{\url{http://ufldl.stanford.edu/housenumbers/}}.
The MNIST dataset is divided into a training set of 60,000 images and a test set of 10,000 images. There are 50,000 training samples and 10,000 testing samples in the CIFAR-10 dataset. In the SVHN dataset, 73,257 digits are used for training and 26,032 digits for testing. These three datasets are all used for the image classification task with 10 categories.

\begin{table}[t]
\centering
\caption{Model architectures.}
\label{tab-local-arch}
%\vspace{-0.1in}
\resizebox{0.47\textwidth}{!}{
\begin{tabulary}{\linewidth}{c|c|c|c}
\toprule
Dataset & ID & \multicolumn{2}{c}{Operation} \\\midrule
\multirow{6}{*}{MNIST}
& & \multicolumn{2}{c}{\textbf{Backbone}}\\
& 1 & \multicolumn{2}{c}{\makecell{Convolution ($10 \times 5 \times 5$) + Max Pooling ($2\times2$)}}\\ 
& 2 & \multicolumn{2}{c}{\makecell{Convolution ($20 \times 5 \times 5$) + Max Pooling ($2\times2$)}} \\ \cline{3-4}
& & \makecell{\textbf{Server-side}}& \makecell{\textbf{Client-side}}\\
& 3 & \makecell{Fully Connected ($320 \times 50$) + ReLU}& \makecell{MLP ($320 \times 320$)} \\ 
& 4 & \makecell{Fully Connected ($50 \times 10$) + Softmax}& \makecell{--} \\ \midrule

% & 3 & \makecell{Fully Connected ($320 \times 50$) + ReLU}& \makecell{MLP ($320 \times 4096$)} \\ 
% & 4 & \makecell{Fully Connected ($50 \times 10$) + Softmax}& \makecell{-} \\ \midrule

\multirow{11}{*}{\makecell{\\\\\\CIFAR \\\& \\ SVHN}}
& & \multicolumn{2}{c}{\textbf{Backbone}}\\
& 1 & \multicolumn{2}{c}{\makecell{Convolution ($32 \times 3 \times 3$) + BatchNorm + ReLU}}\\%\cline{2-3}
& 2 & \multicolumn{2}{c}{\makecell{Convolution ($64 \times 3 \times 3$) + ReLU + Max Pooling ($2\times2$)}}\\
& 3 & \multicolumn{2}{c}{\makecell{Convolution ($128 \times 3 \times 3$) + BatchNorm + ReLU}}\\
& 4 & \multicolumn{2}{c}{\makecell{Convolution ($128 \times 3 \times 3$) + ReLU \\+ Max Pooling ($2\times2$) + dropout(0.05)}}\\
& 5 & \multicolumn{2}{c}{\makecell{Convolution ($256 \times 3 \times 3$) + BatchNorm + ReLU}}\\
& 6 & \multicolumn{2}{c}{\makecell{Convolution ($256 \times 3 \times 3$) + ReLU +Max Pooling ($2\times2$)}}\\
\cline{3-4}
& & \makecell{\textbf{Server-side}}& \makecell{\textbf{Client-side}}\\
& 7 & \makecell{Fully Connected ($4096 \times 1024$) \\+ ReLU + Dropout (0.1)}& \makecell{MLP ($4096 \times 4096$)} \\
& 8 & \makecell{Fully Connected ($1024 \times 512$) \\+ ReLU + Dropout (0.1)}& \makecell{--} \\
& 9 & \makecell{Fully Connected ($512 \times 10$) + Softmax}& \makecell{--} \\

\bottomrule
\end{tabulary}
}
% \vspace{-0.1in}
\end{table}

\begin{table}[h]
\centering
\caption{Parameters chosen in the experiments.}
\label{tab-fl-para}
%\vspace{-0.1in}
\begin{scriptsize}
\begin{tabulary}{\linewidth}{lll}
\toprule
 \textbf{Symbol}&\textbf{Value}&\textbf{Definition}\\
\midrule
$R_G$ & 150(SVHN)/200(others) & the number of global training round\\
$K$ & 100 & the total number of clients\\
$B$ & 10 & the number of active clients\\
$R_L$ & 1 & the number of local epochs \\
$BS_L$ & 10   & the labeled data training batch size \\ 
$BS_U$ & 50   & the unlabeled data training batch size \\ 
$BS_{test}$ & 128   & the testing batch size \\ 
$\gamma$ & 0.01/0.1 & the fraction of labeled data\\
% \midrule
% $lr$ & 0.01   & learning rate \\ 
% $M$ & 0.9   & momentum \\ 
% $wd$ & 1e-4 & weight-decay \\
\bottomrule
\end{tabulary}
%\vspace{-0.1in}
\end{scriptsize}
\end{table}

\smallskip
\noindent\textbf{IID and Non-IID Settings}. In the experiments, each dataset will be randomly shuffled into two parts, i.e., labeled and unlabeled data. 
Given $\gamma$ as the ratio of the labeled data on the entire dataset, there are $|D| * \gamma$ labeled data at the server side, and $|D| * (1-\gamma)$ unlabeled data are distributed to clients, where $|D|$ is the number of training data.  
For the \textbf{IID} setting, both labeled and unlabeled data all have $C$ categories. We equally distribute the unlabeled data to each client, i.e., the number of unlabeled data in each category is the same. 
In the \textbf{Non-IID} setting, the labeled data on the server have all the 10 categories, but each client only contains 2 random categories of unlabeled data. 
We set $\gamma$ as 0.01 or 0.1 in the following experiments.

\smallskip
\noindent\textbf{Baselines}.
To fairly validate the proposed {\name} framework, we use the following state-of-the-art baselines: \textbf{1) FedAvg-FixMatch}~\cite{McMahan2017CommunicationEfficientLO,Sohn2020FixMatchSS}: naive combinations of FedAvg with FixMatch. \textbf{2) FedProx-FixMatch}~\cite{li2018federated,Sohn2020FixMatchSS}:  naive combinations of FedProx with FixMatch. \textbf{3) FedAvg-UDA}~\cite{McMahan2017CommunicationEfficientLO,Xie2019UnsupervisedDA}: naive combinations of FedAvg with UDA. \textbf{4) FedProx-UDA}~\cite{li2018federated,Xie2019UnsupervisedDA}: naive combinations of FedAvg with UDA. \textbf{5) FedMatch}~\cite{Jeong2020FederatedSL}: FedAvg-FixMatch with inter-client consistency and parameter decomposition. \textbf{6) SSFL}~\cite{zhang2020benchmarking}: FedAvg-FixMatch with group normalization (GN) and group averaging (GA).

% \begin{itemize}\scriptsize
%     \item FedAvg-FixMatch~\cite{McMahan2017CommunicationEfficientLO,Sohn2020FixMatchSS}: naive combinations of FedAvg with FixMatch.
%     \item FedProx-FixMatch~\cite{li2018federated,Sohn2020FixMatchSS}:  naive combinations of FedProx with FixMatch.
%     \item FedAvg-UDA~\cite{McMahan2017CommunicationEfficientLO,Xie2019UnsupervisedDA}: naive combinations of FedAvg with UDA.
%     \item FedProx-UDA~\cite{li2018federated,Xie2019UnsupervisedDA}: naive combinations of FedAvg with UDA.
%     \item FedMatch~\cite{Jeong2020FederatedSL}: FedAvg-FixMatch with inter-client consistency \& parameter decomposition.
%     \item SSFL~\cite{zhang2020benchmarking}: FedAvg-FixMatch with group normalization (GN) \& group averaging (GA).
% \end{itemize}

\smallskip
\noindent\textbf{Image augmentations}. {\name} adopts the weak data argumentation technique as in BYOL~\cite{grill2020bootstrap}. First, a random patch of the image is selected and resized to 224 × 224 with a random horizontal flip, followed by a color distortion, consisting of a random sequence of brightness, contrast, saturation, hue adjustments, and an optional grayscale conversion. After that, Gaussian blur and solarization are applied to the patches.

\smallskip
\noindent\textbf{Architecture and Parameters}. We use the same local model for all the baselines and {\name} on each dataset. For the MNIST dataset, we adopt a CNN~\cite{lecun1998gradient} with two 5x5 convolution layers and two linear layers (21,840 parameters in total). For the CIFAR-10 $\&$ SVHN datasets, we apply a CNN with six convolution layers and three linear layers (5,852,170 parameters in total). The details of each model architecture are shown in Table~\ref{tab-local-arch}. Other key parameters used in this paper are shown in Table~\ref{tab-fl-para}.

\begin{table}[t]
\centering
\caption{Average accuracy of three runs on the three datasets under the IID setting with different ratios of labeled data. }
\label{tab:lt}
%\vspace{-0.1in}
\resizebox{0.47\textwidth}{!}{
\begin{tabulary}{\linewidth}{l|ccc|ccc}
\toprule
\multirow{2}{*}{\textbf{Model}}
&\multicolumn{3}{c|}{\textbf{$\gamma=0.01$}}&\multicolumn{3}{c}{\textbf{$\gamma=0.1$}}\\
&{MNIST}&{CIFAR}&{SVHN}&{MNIST}&{CIFAR}&{SVHN}\\
\midrule
FedAvg-FixMatch & 88.67\%   & 49.75\% & 75.05\% &  95.97\%   & 74.75\% & 92.10\%\\
FedProx-FixMatch    & 89.08\%   & 40.58\% & 56.07\% &  96.02\%   & 75.76\% & 91.70\%\\
FedAvg-UDA & 88.67\%   & 43.57\% & 57.98\% &  96.09\%   & 75.32\% & 92.03\%\\
FedProx-UDA    & 88.94\%   & 41.23\% & 65.01\% &  96.40\%   & 76.20\% & 91.65\%\\
FedMatch & 90.16\%   & 52.64\% & 78.52\% &  96.09\%   & 80.16\% & 92.09\%\\
SSFL    & 90.64\%  & 42.34\% & 55.70\% & 96.34\% & 76.00\% & 92.66\%\\
\cline{1-7}
% {\name}  & \textbf{94.79\%}   & \textbf{55.50\%} & \textbf{80.93\%} & \textbf{98.65\%}   & \textbf{80.27\%} & \textbf{92.12\%}\\

{\name}  & \textbf{95.24\%}   & \textbf{54.84\%} & \textbf{81.18\%} & \textbf{98.08\%}   & \textbf{81.47\%} & \textbf{93.19\%}\\
%&{\name}-D   & \textbf{-\%} & -\% & \textbf{-\%} & \textbf{-\%} & \textbf{-\%} & -\%\\
\bottomrule
\end{tabulary}
%\vspace{-0.3in}
}
\end{table}

\subsection{Performance Evaluation for the IID Setting}

We first evaluate {\name}'s performance under the IID setting and report the mean accuracy on three datasets as shown in Table~\ref{tab:lt}. With the ratios of labeled data $\gamma=0.01$, {\name} obtains 95.24\% accuracy on the MNIST dataset (54.84\% in CIFAR-10 and 81.18\% in SVHN), which is a 4.60\% (2.20\% on CIFAR-10 and 2.66\% on SVHN) improvement over the best FedSSL baseline. With $\gamma=0.1$, {\name} still achieves the highest accuracy on all three datasets, but tightens the gap with respect to the classical semi-supervised baselines. This suggests that {\name} receives a significant performance gain when the labeled data in the server are extremely rare. This is because {\name} is built upon the contrastive framework, which has a better capacity to deal with unlabeled data.

We then provide analysis on baselines. FedAvg-FixMatch, FedProx-FixMatch, FedAvg-UDA, and FedProx-UDA are four baselines, which make a simple combination of classical SSL methods (FixMatch and UDA) and FL methods (FedAvg and FedProx). The experiments suggest that FedProx-based methods will receive a better performance than (or close to) FedAvg-based methods when the dataset is simple (MNIST) or the labeled data is plentiful ($\gamma=0.1$). FixMatch-based methods outperform the UDA-based methods when the number of labeled data is extremely limited, and have a close performance to them in the contrary case. FedMatch, making a great improvement on aforementioned methods, achieves the highest performance among all the baselines. SSFL, with its unique group normalization, outperforms the simple combination methods.

\subsection{Performance Evaluation for the Non-IID Settings}

For the non-IID setting, the accuracy of almost all baselines in three datasets is lower than that of IID setting as shown in Table~\ref{tab:ls}, even though they bring more new techniques into the basic SSL-based method. These results illustrate that we need to design an effective way of using unlabeled data while considering the effect of the non-IID setting. Otherwise, the data heterogeneity may degrade the model performance.

From Table~\ref{tab:ls}, we can observe that {\name} outperforms all baselines under the non-IID setting. When $\gamma=0.01$, {\name} obtains 95.67\% accuracy on the MNIST dataset (53.25\% on CIFAR-10 and 81.83\% on SVHN), which is a 3.45\% (0.60\% on CIFAR-10 and 3.99\% on SVHN) improvement over the previous self-supervised state of the art. With $\gamma=0.1$, {\name} still achieves the highest accuracy on all three datasets with an obvious performance gain. We can conclude that the contrastive network designed in {\name} has more advantages for dealing with the non-IID setting than any other baselines.

\begin{table}[t]
\centering
\caption{Average accuracy of three runs on the three datasets under the non-IID setting with different ratios of labeled data. }
\label{tab:ls}
%\vspace{-0.1in}
\resizebox{0.47\textwidth}{!}{
\begin{tabulary}{\linewidth}{l|ccc|ccc}
\toprule
\multirow{2}{*}{\textbf{Model}}
&\multicolumn{3}{c|}{\textbf{$\gamma=0.01$}}&\multicolumn{3}{c}{\textbf{$\gamma=0.1$}}\\
&{MNIST}&{CIFAR}&{SVHN}&{MNIST}&{CIFAR}&{SVHN}\\
\midrule
FedAvg-FixMatch & 89.27\%   & 48.93\% & 75.92\% &  96.15\%   & 75.88\% & 92.08\%\\
FedProx-FixMatch    & 91.98\%   & 42.08\% & 62.67\% & 95.89\%   & 75.36\% & 92.05\%\\
FedAvg-UDA & 89.98\%   & 42.07\% & 56.64\% &  96.21\%   & 75.31\% & 92.04\%\\
FedProx-UDA    & 88.93\%   & 43.66\% & 62.50\% &  95.59\%   & 75.66\% & 92.26\%\\
FedMatch & 89.61\%   & 52.65\% & 77.84\% &  95.70\%   & 79.50\% & 91.46\%\\
SSFL    & 92.22\%  & 43.27\% & 67.33\% & 96.57\% & 75.16\% & 92.21\%\\
\cline{1-7}
{\name}  & \textbf{95.67\%}   & \textbf{53.25\%} & \textbf{81.83\%} & \textbf{98.22\%}   & \textbf{81.96\%} & \textbf{92.67\%}\\
%&{\name}-D   & \textbf{-\%} & -\% & \textbf{-\%} & \textbf{-\%} & \textbf{-\%} & -\%\\
\bottomrule
\end{tabulary}

}
\end{table}

\subsection{Ablation Studies}

\begin{figure*}[t]
\centering
%\hspace*{-1.1cm} 
\includegraphics[width=0.9\textwidth]{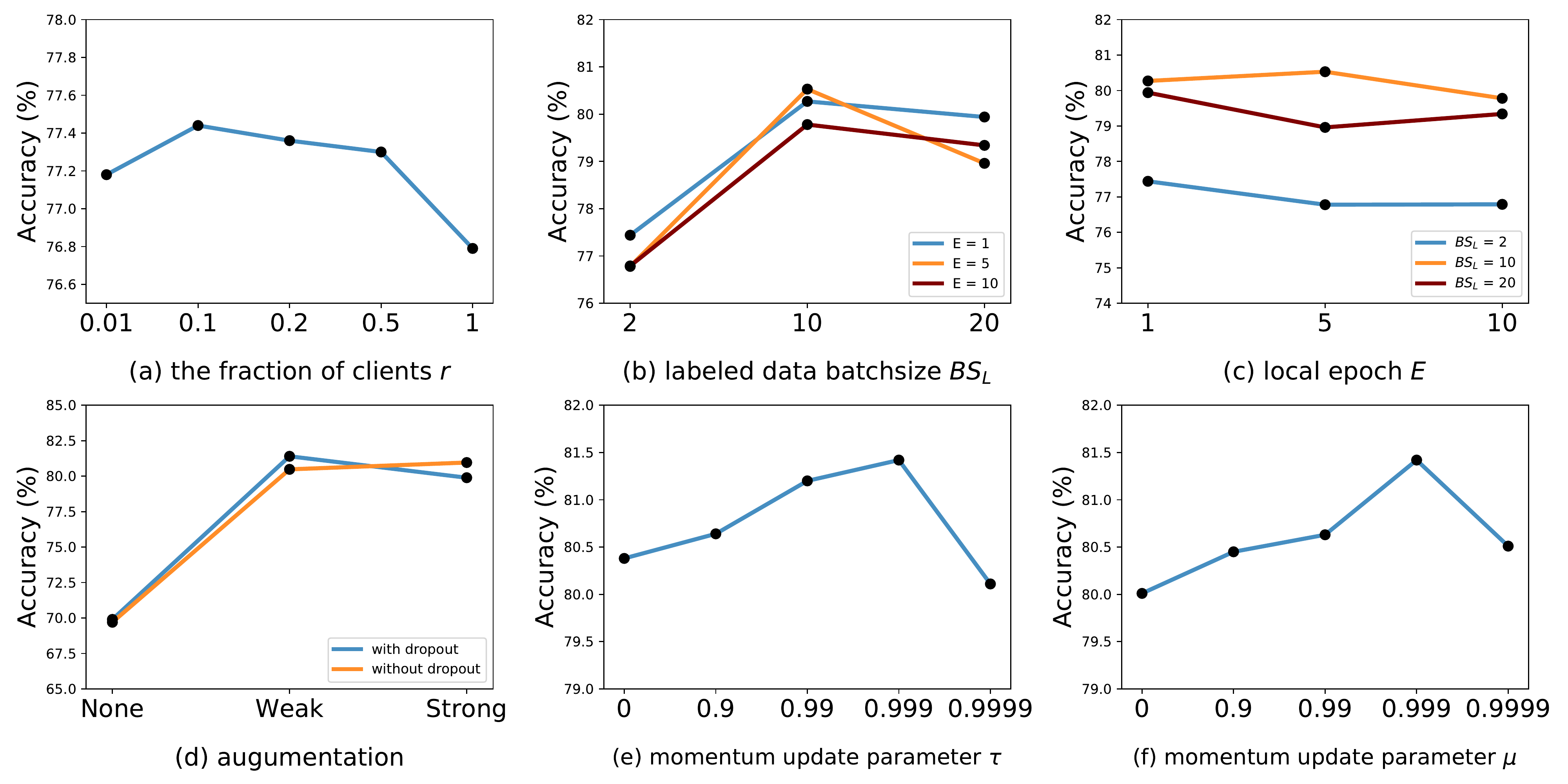} % Reduce the figure size so that it is slightly narrower than the column.
\caption{Mean accuracy on CIFAR-10 with $\gamma=0.1$ over four runs per hyperparameter setting. In each experiment, we vary one hyperparameter and fix other hyperparameters listed in Table~\ref{tab-fl-para}.}
\label{ablation}
% \vspace{-0.2in}
\end{figure*}

To assess the importance of various aspects of the model, we conduct experiments on CIFAR-10 with $\gamma =0.1$, varying one or a few hyperparameters at a time while keeping the others fixed.

\smallskip
\noindent\textbf{Client fraction $r$ (Figure~\ref{ablation}(a))}. We firstly experiment with the client fraction $r$, which controls the amount of multi-client parallelism. Specifically, the client fraction $r$ is defined as the fraction of the number of chosen clients $B$ among all clients $K$, i.e., $\frac{B}{K}$. We report the accuracy value with total 200 rounds training with the IID case. With $BS_L = 10$, there is no significant advantage in increasing the client fraction. This result demonstrates that the number of training unlabeled data has limited influence on the final performance, and the number of labeled data at the server domains the performance in our FedSSL setting. These results suggest us setting $r = 0.1$ for the following experiments, which strikes a good balance between computational efficiency and convergence rate.

\smallskip
\noindent\textbf{Batch size $BS_L$ (Figure~\ref{ablation}(b)) \& Local epoch $E$ (Figure~\ref{ablation}(c))}. In this section, we fix $r = 0.1$ and add more computation per client on each round, either decreasing $BS_L$, increasing $E$, or both. Note that we fix $BS_U \triangleq 5*BS_L$ to have a precise measurement of batch size. We firstly validate the model performance by test accuracy in three different batch size $BS_L$ settings. In the real-world applications, the batch size $BS_L$ has a close relationship to the client hardware. As $BS_L$ is getting larger to take full advantage of available parallelism on the client hardware, the computation time will reduce dramatically in each client, which leads to higher efficiency. As the value of batch size $BS_L$ grows, the test accuracy firstly increases rapidly before reaching the peak at $BS_L=10$ and decreases slowly as the batch size keeps growing. Based on these observations, we fix $BS_L = 10$ in the experiments, which strikes a good balance between model performance and computation time. 

Previous work~\cite{McMahan2017CommunicationEfficientLO} suggests that adding more local SGD updates per round, i.e., increasing local epoch $E$, can produce a dramatic decrease in communication costs when we fix the sum of local epoch. We conduct experiments to evaluate the impact of local epoch $E$ under different batch size $BS_L$ settings. As shown in Figure~\ref{ablation}(c), the test accuracy for 200 communication rounds is relatively stable when local epoch $E$ increases. Based on these results, we concludes that local epoch $E$ has limited influence on model training in FedSSL, which is similar to the classical FL settings~\cite{McMahan2017CommunicationEfficientLO}. We choose $E=1$ for most of our experiments.

\smallskip
\noindent\textbf{Data augmentation \& Dropout (Figure~\ref{ablation}(d))}. To confine the best augmentation technique under the FedSSL scenario, we conduct the ablation experiments on three different data augmentation strategy (None, Weak as BYOL~\cite{grill2020bootstrap}, Strong as FixMatch~\cite{Sohn2020FixMatchSS}) and two settings (with or without dropout). We can observe that the model with weak augmentation on input images and using dropout receives the best accuracy on the test set, while the model with none or strong augmentation has a significant performance decrease. On the other hand, dropout has limited benefits when augmentation is absent. Without dropout, the strong augmentation model performs the best among other settings, but only has a small gap between the weak augmentation version. Based on these results, for most of our experiments, we choose weak augmentation and dropout as the model noise, which receives a good performance among all the settings.

\smallskip
\noindent\textbf{Momentum update (Figure~\ref{ablation}(e) and~\ref{ablation}(f))}. Two essential hyperparameters of {\name} are the EMA decay on server training $\tau$ and local $\mu$ training. We conduct experiments to select the best value of $\tau$ and $\mu$ and validate the sensitivity of our model to these values. We can see that in each case the good values span roughly an order of magnitude and outside these ranges the performance degrades quickly. Note that we use the ramp-up technique with an upper EMA decay during training. We choose this strategy because the online net improves quickly early in the training, and thus the target net should forget the old, inaccurate, online weights quickly. According to these results, we use $\tau=0.999$ and $\mu = 0.999$ in each training run, which receives the best performance among all the settings.

\begin{figure}[t]
\centering
\includegraphics[width=0.4\textwidth]{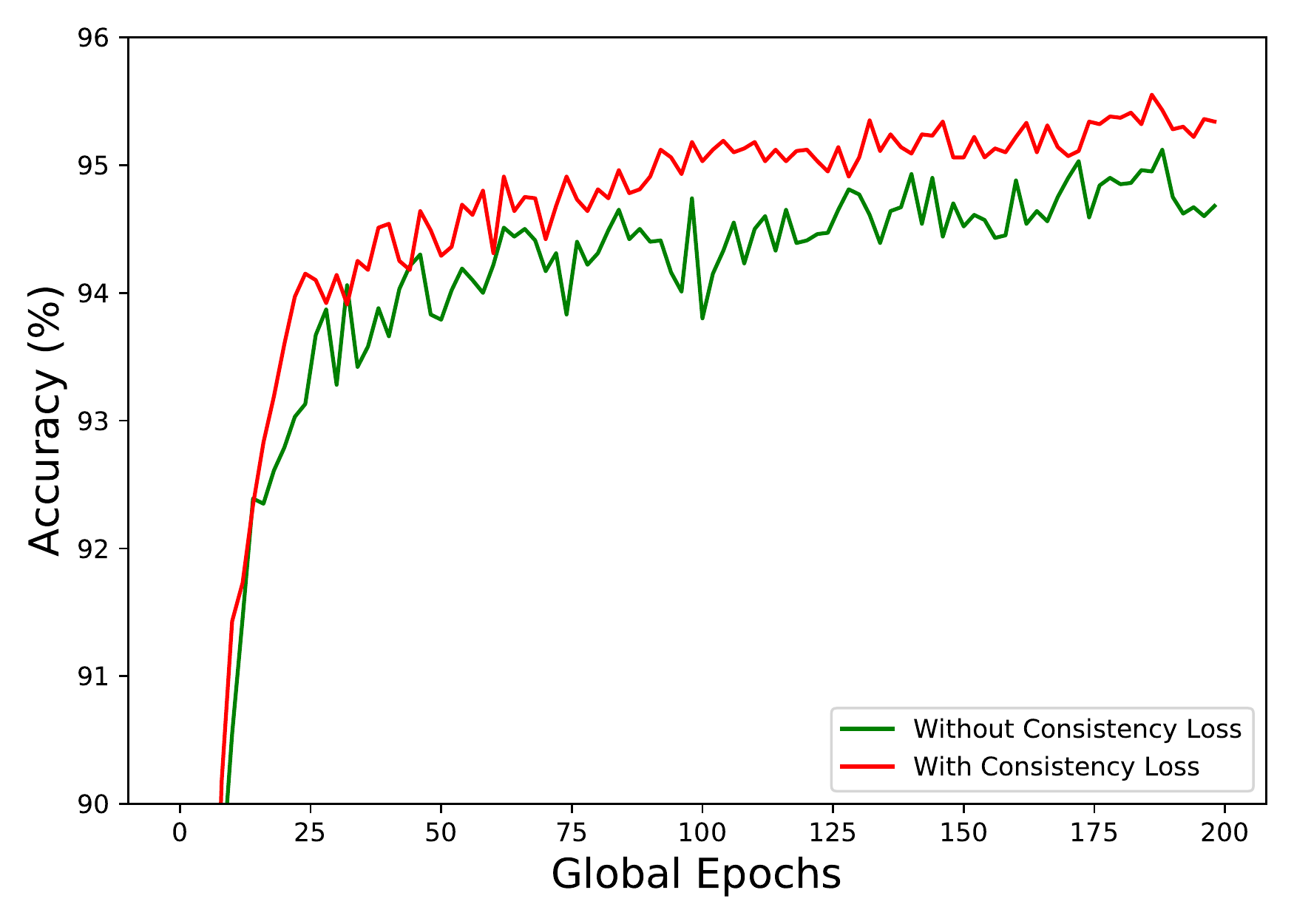} % Reduce the figure size so that it is slightly narrower than the column.
%\vspace{-0.2in}
\caption{Mean accuracy on MNIST ($\gamma=0.01$ and $\beta=0.05$) over four runs per hyperparameter setting, where $\gamma$ represents the percentage of labeled data at server, and $\beta$ means the percentage of unlabeled data at server.}
\label{consistency}
%\vspace{-0.1in}
\end{figure}

\subsection{Importance of Consistency Loss for Unlabeled Server Data}\label{consostency_loss}
In this section, we conduct a preliminary experiment to validate the performance of {\name} in a more general scenario, where the server has both the labeled data and the unlabeled data. The designed consistency loss in Eq.~(\ref{mse_loss}) is able to extract information from the unlabeled data in the server. 
%Therefore, we validate our model under a new setting, where the server has both the labeled data and the unlabeled data. 
Ideally, injecting unlabeled data to the model is able to increase the performance.
We report the test-set accuracy and accuracy curve with total 200 rounds training in the IID case. From Figure~\ref{consistency}, we can observe that the model with consistency loss receives a significant performance increase during the training. When $\gamma=0.01$ and $\beta=0.05$, {\name} with consistency loss obtains 95.55\% accuracy on MNIST dataset, compared with the one without consistency loss obtains only 94.68\% accuracy. This result demonstrates that the consistency loss in Eq.~(\ref{mse_loss}) has a great influence on the final performance when the server has both the labeled data and unlabeled data. 
% \begin{figure}[!h]
% \centering
% \includegraphics[width=\textwidth]{pic/curve.jpg} % Reduce the figure size so that it is slightly narrower than the column.
% \caption{An example for illustrating three Non-IID settings.} 
% \label{curve}
% \end{figure}
\section{Related Work}
% \jq{One possible suggestion: should we talk about federated semi-supervised learning after the federated learning and semi-supervised learning?}

\noindent\textbf{Federated learning} (FL) aims to collaboratively train a joint model using data from different parties or clients. Most algorithms of FL focus on the supervised setting and mainly solve three challenges: statistical heterogeneity~\cite{DBLP:journals/corr/abs-1806-00582,DBLP:journals/corr/abs-1812-06127,DBLP:journals/corr/abs-1811-12629}, system constraints~\cite{Caldas2018ExpandingTR,Wang2019CMFLMC,Chen2018FederatedMW}, and trustworthiness~\cite{Bhowmick2018ProtectionAR,Geyer2017DifferentiallyPF,Bonawitz2016PracticalSA}. In this paper, we only discuss the challenge of statistical heterogeneity, i.e., the Non-IID setting. To address this challenge, various algorithms have been proposed like sharing a some part of data~\cite{DBLP:journals/corr/abs-1806-00582}, training personal model for each client~\cite{DBLP:journals/corr/abs-1812-06127}, or adjusting the SGD convergence of FL \cite{DBLP:journals/corr/abs-1811-12629}. However, relative little attention has been paid to solve the data heterogeneity in federated semi-supervised learning~\cite{Albaseer2020ExploitingUD}.

\smallskip
\noindent\textbf{Federated semi-supervised learning} (FedSSL), which introduces unlabeled data into federated learning, significantly increases the difficulty of the analysis of model training. Several approaches are proposed to tackle the FedSSL problem by integrating classical semi-supervised learning into the federated learning framework, such as FedSem~\cite{Albaseer2020ExploitingUD}, FedMatch~\cite{Jeong2020FederatedSL}, and SSFL~\cite{zhang2020benchmarking}. FedSem~\cite{Albaseer2020ExploitingUD} is a simple two-phase training with pseudo labeling. A study on inter-client consistency suggests that a simple application of SSL methods might not perform well in FL, and the inter-client level consistency might improve the performance~\cite{Jeong2020FederatedSL}. The group normalization (GN) and group averaging (GA) techniques are proposed to decrease gradient diversity and further improve the Non-IID problem. However, these efforts do not propose a more general and practical algorithm and validate their potentials on new challenges of federated semi-supervised learning.

\smallskip
\noindent\textbf{Semi-supervised learning} (SSL) mitigates the requirement for labeled data by providing a way of leveraging unlabeled data~\cite{4787647}. 
%Since unlabeled data can often be obtained with minimal human labor, any performance boost conferred by SSL often comes with low cost. 
The recent studies in SSL are diverse but one trend of unity. Pseudo labeling, which converts unlabeled data to labeled data, utilizes unlabeled data by labeling the data with a dynamic threshold~\cite{Lee2013PseudoLabelT}. A nature and well-working idea on consistency regularization has been widely adopted in SSL~\cite{Rasmus2015SemisupervisedLW,Tarvainen2017MeanTA,Laine2017TemporalEF,Miyato2019VirtualAT,park2017adversarial}. A further discussion on how loss geometry interacts with training procedures suggests that the flat platform of SGD leads to the convergence dilemma of consistency-based SSL~\cite{Athiwaratkun2019ThereAM}. By exploring further or mixing many practical methods, UDA~\cite{Xie2019UnsupervisedDA}, MixMatch~\cite{Berthelot2019MixMatchAH}, ReMixMatch~\cite{Berthelot2019ReMixMatchSL}, and Fixmatch~\cite{Sohn2020FixMatchSS} are proposed. In our work, we mainly focus on utilizing the pure consistency-based methods working with federated learning.
\section{Conclusion}
In this work, we focus on a practical and challenging setting in federated semi-supervised learning (FedSSL), i.e., all the labeled data are stored in the sever and unlabeled data are in the clients. To fully consider the new fundamental challenges caused by unlabeled data, we propose a novel yet general framework, called {\name}, which is not only effective and robust for several real-world FedSSL scenarios but also takes data heterogeneity into consideration. Experiments on three image datasets under both IID and non-IID settings demonstrate the effectiveness of the proposed {\name} framework compared with state-of-the-art baselines for the federated semi-supervised learning task. Moreover, we conduct ablation experiments to analyze the insights and illustrate the generality and characteristic of our model.

\bibliographystyle{IEEEtran}
\bibliography{references.bib}

\end{document}